\documentclass[sigconf,sigconf]{acmart}
\usepackage{makecell}
\usepackage{booktabs}

\AtBeginDocument{%
  \providecommand\BibTeX{{%
    \normalfont B\kern-0.5em{\scshape i\kern-0.25em b}\kern-0.8em\TeX}}}

\setcopyright{acmcopyright}
\copyrightyear{2023}
\acmYear{2023}
\acmDOI{XXXXXXX.XXXXXXX}

\acmConference[Conference acronym 'XX]{The 21st ACM International Conference on Mobile Systems, Applications, and Services}{June 18--12, 2023}{Helsinki, Finland}
\acmPrice{15.00}
\acmISBN{978-1-4503-XXXX-X/18/06}

\copyrightyear{2023} 
\acmYear{2023} 
\setcopyright{rightsretained} 
\acmConference[MobiSys '23]{The 21st Annual International Conference on Mobile Systems, Applications and Services}{June 18--22, 2023}{Helsinki, Finland}
\acmBooktitle{The 21st Annual International Conference on Mobile Systems, Applications and Services (MobiSys '23), June 18--22, 2023, Helsinki, Finland}\acmDOI{10.1145/3581791.3597371}
\acmISBN{979-8-4007-0110-8/23/06}

\begin{document}

\title{Poster: Towards Battery-Free Machine Learning Inference and Model Personalization on MCUs}


\author{Yushan Huang}
\affiliation{%
  \institution{Imperial College London}
  \city{London}
  \country{UK}}
\email{yushan.huang21@imperial.ac.uk}

\author{Hamed Haddadi}
\affiliation{%
  \institution{Imperial College London}
  \city{London}
  \country{UK}}
\email{h.haddadi@imperial.ac.uk}

\renewcommand{\shortauthors}{Yushan and Hamed, et al.}


\begin{abstract}

Machine learning (ML) is moving towards edge devices. However, ML models with high computational demands and energy consumption pose challenges for ML inference in resource-constrained environments, such as the deep sea. To address these challenges, we propose a battery-free ML inference and model personalization pipeline for microcontroller units (MCUs). As an example, we performed fish image recognition in the ocean. We evaluated and compared the accuracy, runtime, power, and energy consumption of the model before and after optimization. The results demonstrate that, our pipeline can achieve 97.78\% accuracy with 483.82 $\mathbf{KB}$ Flash, 70.32 $\mathbf{KB}$ RAM, 118 $\mathbf{ms}$ runtime, 4.83 $\mathbf{mW}$ power, and 0.57 $\mathbf{mJ}$ energy consumption on MCUs, reducing by 64.17\%, 12.31\%, 52.42\%, 63.74\%, and 82.67\%, compared to the baseline. The results indicate the feasibility of battery-free ML inference on MCUs.

\end{abstract}





\maketitle

\section{Introduction}
\label{sec:intro}


Machine learning (ML) models have become ubiquitous in solving diverse problems. However, these models often demand high memory, computation power, and energy requirements, posing challenges for ML deployment on resource-constrained edge devices. The edge offers several advantages such as reduced response latency, better bandwidth utilization, and improved security and privacy expectations. Therefore, there is a pressing need to develop optimized lightweight ML models for deployment on the edge.

To address this problem, researchers have explored model compression techniques to compress the model for the edge \cite{niu2020patdnn}. However, these technologies typically assume that the edge has sufficient memory, computing power, and power supply, which can be challenging to achieve in extreme environments such as deep seas, and remote areas. Some studies have deployed traditional models such as SVM on MCUs \cite{sudharsan2020edge2train}, but they require manual feature extraction and may not perform well on high-dimensional data. Additionally, power and energy consumption are often overlooked in edge-based ML deployment. Recent advancements in battery-free sensing technology have allowed for innovative applications \cite{guida2020underwater}. These techniques have made long-term MCUs use possible without the need for power supplies or batteries. this study aims to examine the feasibility of achieving battery-free ML inference and model personalization on MCUs for extreme environments.


In our previous work, we have achieved battery-free inference for sound signal classification \cite{10.1145/3508396.3512877}. However, we did not consider optimizing the model and energy consumption. The primary contributions of this paper are as follows:

(1) This pipeline is an end-to-end solution designed for the efficient design, deployment, energy-optimizing, and execution of ML models on resource-constrained MCUs.

(2) We optimized the Proxyless neural architecture search (ProxylessNAS) \cite{cai2018proxylessnas} model, resulting in a reduction of the model size from 1350.25 $\mathbf{KB}$ Flash and 80.20 $\mathbf{KB}$ RAM to 483.82 $\mathbf{KB}$ Flash and 70.32 $\mathbf{KB}$ RAM, with a slight loss of approximately 0.1\% in accuracy.

(3) We compared the runtime, power, and energy consumption of the model before and after optimization by Monsoon Power Monitor \cite{msoon}. Compared with the unoptimized model, we achieved 97.78\% accuracy with a runtime of 118 $\mathbf{ms}$, power of 4.83 $\mathbf{mW}$, and energy consumption of 0.57 $\mathbf{mJ}$, which reduced by 52.42\%, 63.74\%, and 82.67\%, respectively.

\begin{figure}[t]
\vskip 0.2in
\begin{center}
\centerline{\includegraphics[width=0.4\textwidth]{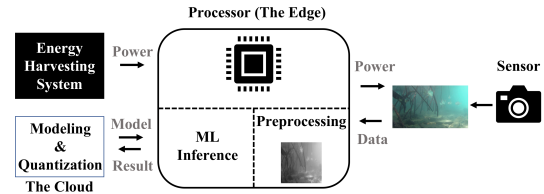}}
\caption{Design of battery-free inference on MCUs}
\label{fig:pipeline}
\end{center}
\vskip -0.2in
\end{figure}

\begin{figure*}[t]
\vskip 0.2in
\begin{center}
\centerline{\includegraphics[width=0.85\textwidth]{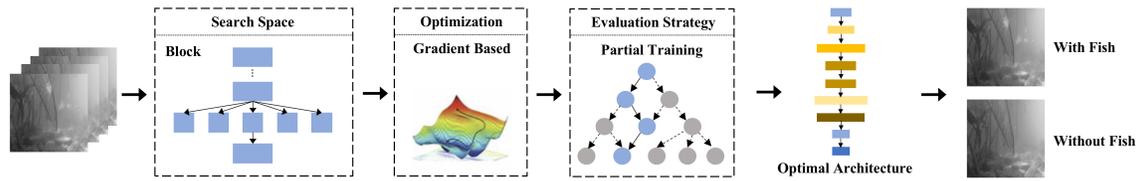}}
\caption{The main components of ProxylessNAS. }
\label{fig:proxylessnass}
\end{center}
\vskip -0.2in
\end{figure*}

\section{Pipeline}
\label{sec:pipeline}

The pipeline mainly has four components: preprocessing, modeling, quantization, and deployment and inference, as shown in Fig.~\ref{fig:pipeline}.

\textbf{Preprocessing}
We conducted a study on fish recognition in the ocean using DeepFish dataset \cite{saleh2020realistic}. DeepFish is a dataset based on field records of fish habitats and customized for the analysis of fish in underwater marine environments. The dataset comprises approximately 40,000 high-resolution 1,920$\times$1,080 data points. As Fig.~\ref{fig:pipeline} (a) shows, the original images with three RGB channels at a resolution of 1,920$\times$1,080 are reconstructed into single RGB channel images at a resolution of 32$\times$32.

\textbf{Modeling.} We optimized the ProxylessNAS \cite{cai2018proxylessnas} to search the approximate model. ProxylessNAS aims to directly search for architectures for target tasks, and optimizes model weights and architecture parameters alternatively using gradient-based methods. ProxylessNAS introduces binary gates that binarize the architecture parameters of an overparameterized network. This enables only one path to load at runtime, reducing memory consumption by not loading the entire overparameterized network to update model weights. The main components of ProxylessNas are shown in Fig.~\ref{fig:proxylessnass}.

\textbf{Quantization.} To reduce the model's size, computation requirements, power, and energy consumption, we utilize static quantization to quantize the original model. Static quantization is an optimization technique for neural network models, which converts the parameters and activations from floating-point to integer representations while preserving the model's accuracy. The process involves data collection, quantization range determination, quantization conversion, and dequantization. Static quantization significantly reduces model size, speeds up inference, and lowers power and energy consumption while maintaining model accuracy.



\textbf{Deployment.} We utilize X-CUBE-AI to deploy the model and inference process into \emph{.h} and \emph{.c} files. X-CUBE-AI is a software package that helps developers integrate models into embedded applications on MCUs. It includes tools such as a neural network model converter and inference engine. 
\section{Evaluation}
\label{sec:evaluation}


We trained the lightweight ProxylessNAS model by TensorFlow, transferred it to TFLite, and applied static quantization to reduce the model size, computation requirement, power, and energy consumption. We evaluated and compared the offline accuracy, and conducted 10 repeated experiments on the original TensorFlow model, original TFLite model, and optimized TFLite model. The accuracies of these models are 97.88±1.12\%, 97.88±1.12\%, and 97.78±1.08\%. The quantized TFLite Model only loses approximately 0.1\% accuracy. 


We transplanted the TFLite models to the STM32L4R5 development kit, and utilized the Monsoon power monitor to measure power and energy. Power was calculated as the product of the current and voltage when the board was connected to the power source. The board was supplied power at 1.9$\mathbf{V}$. The power consumption results are presented in Table.~\ref{tab:modelpower}. Previous research has shown that underwater acoustic and ultrasonic signals can generate a few $\mathbf{mW}$ \cite{guida2020underwater}, suggesting that our pipeline can run solely on harvested energy.

\begin{table}[t]
\centering
\caption{Evaluation of the power performance on MCUs}
\label{tab:modelpower}
\begin{tabular}{@{}ccccccc@{}}
\toprule
           & \thead{\fontsize{10}{10}\selectfont Flash\\\fontsize{10}{10}\selectfont$(\mathbf{KB})$} & \thead{\fontsize{10}{10}\selectfont RAM\\\fontsize{10}{10}\selectfont$(\mathbf{KB})$} & \thead{\fontsize{10}{10}\selectfont Power\\\fontsize{10}{10}\selectfont $( \mathbf{mW})$} & \thead{\fontsize{10}{10}\selectfont Time\\\fontsize{10}{10}\selectfont $(\mathbf{ms})$} & \thead{\fontsize{10}{10}\selectfont Energy\\\fontsize{10}{10}\selectfont $(\mathbf{mJ})$} \\ \midrule
Original Model   & 1350.25  & 80.20 & 13.32 & 248 & 3.29\\ 
Optimized Model      & 483.82  & 70.32  & 4.83 & 118 & 0.57\\ 
\bottomrule
\end{tabular}
\end{table}

\section{Conclusion and Future Work}

This paper introduces an energy-optimized ML deployment pipeline for resource-constrained MCUs. Compared to the unoptimized model, we achieved an average accuracy of 97.78\% with 483.82 $\mathbf{KB}$ Flash, 70.32 $\mathbf{KB}$ RAM, 118 $\mathbf{ms}$ inference time, 4.83 $\mathbf{mW}$ power, and 0.57 $\mathbf{mJ}$ energy consumption, which reduced by 64.17\%, 12.31\%, 52.42\%, 63.74\%, and 82.67\%, respectively. The results indicate the viability of battery-free ML on MCUs, with the potential to harvest energy from certain devices. In the future, we plan to further optimize energy consumption and improve the personalization of the model to address the data heterogeneity problem. We will also test the pipeline on various models and tasks to assess its scalability.

\label{sec:conclusion}

\bibliographystyle{unsrt}

\bibliography{sample-base}

\end{document}